\documentclass[10pt,twocolumn,letterpaper]{article}

\usepackage[pagenumbers]{cvpr} 

\usepackage{colortbl}
\usepackage{multirow}
\usepackage[accsupp]{axessibility}

\definecolor{cvprblue}{rgb}{0.21,0.49,0.74}
\usepackage[pagebackref,breaklinks,colorlinks,allcolors=cvprblue]{hyperref}

\def\paperID{5413}
\def\confName{CVPR}
\def\confYear{2025}

\title{GaussTR: Foundation Model-Aligned Gaussian Transformer for \\ Self-Supervised 3D Spatial Understanding}

\author{
    Haoyi Jiang$^{1,}$\thanks{Intern at Horizon Robotics: \texttt{haoyi\_jiang@hust.edu.cn}} \quad Liu Liu$^{2}$ \quad Tianheng Cheng$^{1}$ \quad Xinjie Wang$^{2}$ \quad Tianwei Lin$^{2}$ \\
    Zhizhong Su$^{2}$ \quad Wenyu Liu$^{1}$ \quad Xinggang Wang$^{1,}$\thanks{Corresponding author: \texttt{xgwang@hust.edu.cn}} \\
    $^1$Huazhong University of Science \& Technology \quad $^2$Horizon Robotics \\
}

\begin{document}
\maketitle
\begin{abstract}

3D Semantic Occupancy Prediction is fundamental for spatial understanding, yet existing approaches face challenges in scalability and generalization due to their reliance on extensive labeled data and computationally intensive voxel-wise representations.
In this paper, we introduce \textbf{GaussTR}, a novel \textbf{Gauss}ian-based \textbf{TR}ansformer framework that unifies sparse 3D modeling with foundation model alignment through Gaussian representations to advance 3D spatial understanding. GaussTR predicts sparse sets of Gaussians in a feed-forward manner to represent 3D scenes.
By splatting the Gaussians into 2D views and aligning the rendered features with foundation models, GaussTR facilitates self-supervised 3D representation learning and enables open-vocabulary semantic occupancy prediction without requiring explicit annotations.
Empirical experiments on the Occ3D-nuScenes dataset demonstrate GaussTR's state-of-the-art zero-shot performance of 12.27 mIoU, along with a 40\% reduction in training time. These results highlight the efficacy of GaussTR for scalable and holistic 3D spatial understanding, with promising implications in autonomous driving and embodied agents. The code is available at \url{https://github.com/hustvl/GaussTR}.

\end{abstract}
\section{Introduction}
\label{sec:intro}

\begin{figure}
    \centering
    \includegraphics[width=0.475\textwidth]{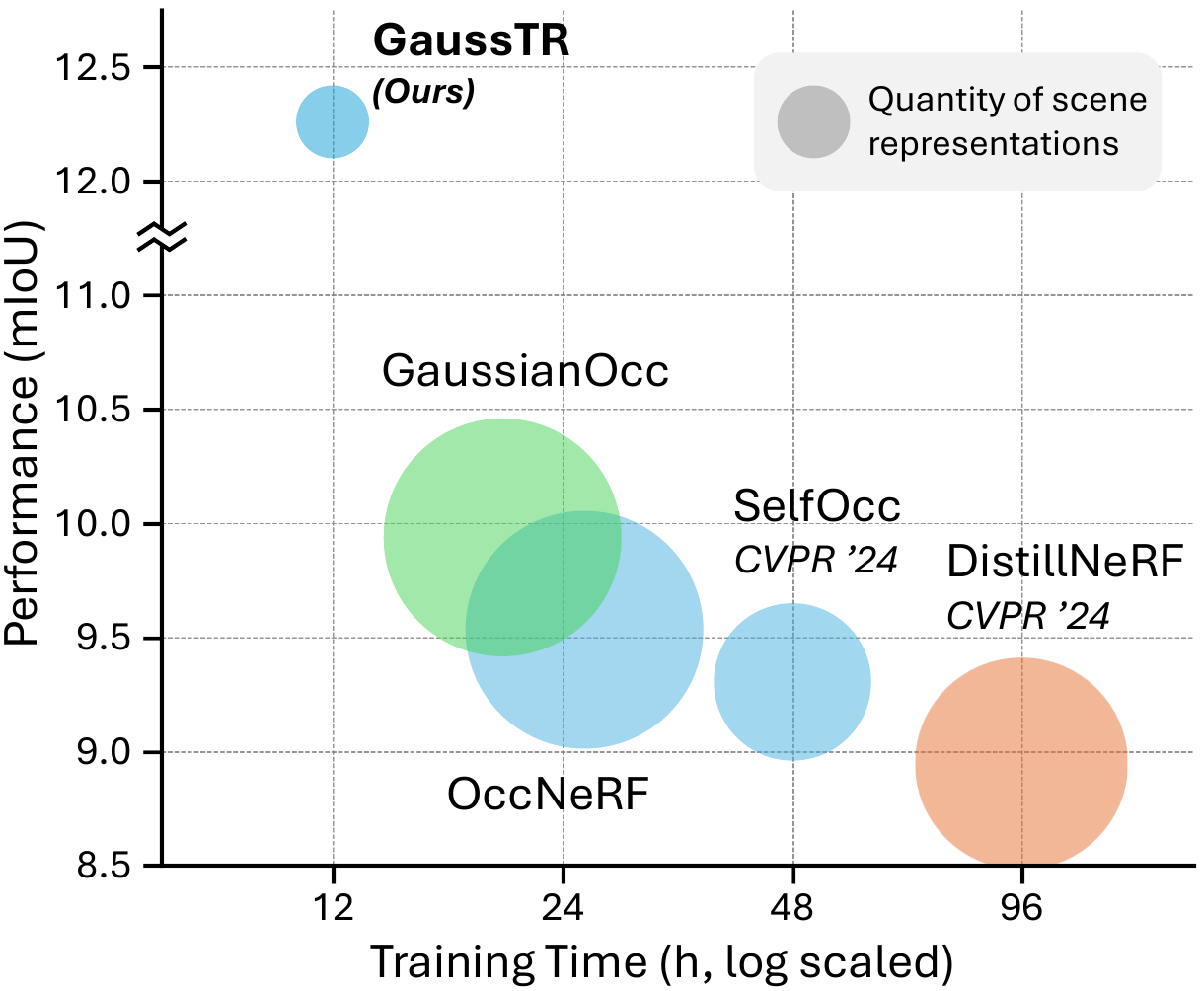}
    \caption{\textbf{Comparative performance of self-supervised 3D occupancy prediction methods.} GaussTR achieves a 2.33 mIoU (23\%) improvement over other counterparts while reducing training time by approximately 40\%, using merely 3\% of the scene representation parameters (\eg, voxels or Gaussians). Marker sizes are proportional to the logarithm of scene representation parameters.}
    \label{fig:performance_comp}
\end{figure}

The pursuit of artificial general intelligence has underscored 3D spatial understanding as a cornerstone \textit{en route to} higher levels of autonomy, empowering autonomous systems to perceive, reason about, and interact with complex 3D environments for applications in autonomous driving and embodied agents. While Vision Foundation Models (VFMs)—such as CLIP~\cite{CLIP}, DINO~\cite{DINO, DINOv2}, and EVA~\cite{EVA, EVA02}—have pushed the boundaries of 2D visual perception, their extension to self-supervised 3D spatial understanding remains relatively uncharted due to the inherent gap between 2D vision and 3D spatial intelligence.

3D Semantic Occupancy Prediction aims to generate comprehensive spatial cognition by predicting voxel-wise occupancy and semantics, forming a fundamental task for spatial understanding. While prevalent supervised methods~\cite{TPVFormer, VoxFormer, OccFormer, Symphonies} have attained superior performance, they heavily depend on extensive 3D annotations, which are costly and labor-intensive to acquire. Besides, dense voxel representations impose substantial computational overhead and struggle with capturing high-level contextual relationships attributed to excessive granularity. Recent self-supervised approaches~\cite{SelfOcc, OccNeRF, DistillNeRF, GaussianOcc}, inspired by RenderOcc~\cite{RenderOcc}, have sought to alleviate the data constraints by leveraging temporal image reprojection consistency~\cite{MonoDepth2} to provide geometric supervision. However, their reliance on pre-computed 2D segmentation pseudo-labels restricts their ability to learn generalizable representations, leaving them prone to out-of-distribution scenarios in real-world applications.

Drawing inspiration from the emerging 3D Gaussian splatting (GS)~\cite{3DGS} for scene reconstruction, we propose \textbf{GaussTR}, a novel \textbf{Gauss}ian-based \textbf{TR}ansformer that integrates sparse 3D modeling with foundation models through Gaussian representations to facilitate self-supervised 3D spatial understanding. GaussTR models scenes as sparse, unstructured sets of Gaussians, predicted through a series of Transformer~\cite{Transformer} layers in a feed-forward manner. Specifically, GaussTR aggregates multi-view VFM features for individual Gaussians with deformable cross-attention~\cite{DeformableDETR}, followed by global self-attention across Gaussian queries for effective 3D modeling. The resultant Gaussian parameters are regressed via MLPs within a dedicated Gaussian head.

Exploiting the consistency of Gaussian splatting across 2D and 3D modalities, GaussTR renders Gaussians back to 2D views and enforces feature alignment with foundation models. GaussTR thus learns versatile 3D representations enriched with broad visual priors, enabling open-vocabulary occupancy prediction by similarity measurement with target categories without explicit annotations. Optionally, auxiliary segmentation supervision from Grounded SAM~\cite{SAM, GroundedSAM} is incorporated to further refine rendered Gaussian boundaries.

Extensive experiments on the Occ3D-nuScenes~\cite{Occ3D} dataset demonstrate that GaussTR achieves state-of-the-art zero-shot performance of 12.27 mIoU—a 23\% improvement over previous methods—while simultaneously reducing training time by 40\%, as shown in \cref{fig:performance_comp}. Ablation studies further validate the efficacy of our design choices. These results underscore the synergistic benefits of our proposed Gaussian-based 3D modeling and foundation model alignment, charting a scalable pathway for generalizable 3D spatial understanding.

In summary, our contributions are as follows:

\begin{itemize}
    \item \textbf{\textit{Scene Representation as Sparse Gaussians.}} GaussTR introduces a novel Gaussian-based 3D modeling method through feed-forward Transformer prediction, eliminating dense voxel grids and alleviating computational overhead.
    \item \textbf{\textit{Foundation Model Alignment for Self-Supervised Learning.}} GaussTR learns generalizable 3D representations aligned with foundation models via differentiable splatting, facilitating self-supervised open-vocabulary occupancy prediction without requiring explicit labels.
    \item \textbf{\textit{State-of-the-Art Zero-Shot Performance.}} GaussTR achieves 12.27 mIoU on Occ3D-nuScenes, outperforming previous methods by 23\% alongside a 40\% reduction in training time, underscoring the efficacy of Gaussian-based 3D modeling and foundation model alignment.
\end{itemize}

\section{Related Works}
\label{sec:related_works}

\subsection{3D Semantic Occupancy Prediction}

3D Semantic Occupancy Prediction, formerly known as 3D Semantic Scene Completion (SSC)~\cite{SSCNet}, aims to predict the occupancy and semantics for each voxel grid within a 3D scene volume. By providing spatial awareness, 3D occupancy is crucial for applications such as navigation and obstacle avoidance in the fields of autonomous driving and embodied agents. Prevalent approaches have explored 2D-to-3D projection mechanisms~\cite{MonoScene, VoxFormer, BEVFormer, BEVDet}, efficient scene representations~\cite{TPVFormer, Symphonies, SparseOcc, GaussianFormer, OSP, FlashOcc}, and architectural advancements~\cite{OccFormer, VoxFormer, Occ3D, Cam4DOcc, COTR, BVT}. Despite their contributions, these supervised methods depend on detailed voxel-wise annotations, which are costly to obtain and constrain their scalability.

RenderOcc~\cite{RenderOcc} alleviates this constraint by leveraging volume rendering~\cite{NeRF} to derive 3D supervision exclusively from 2D semantic and depth labels. Following RenderOcc, several self-supervised methods~\cite{SelfOcc, OccNeRF, S4C} have been proposed, treating occupancy volumes as radiance fields and using the temporal consistency of rendered or reprojected images~\cite{MonoDepth2, BTS, SceneRF} to provide geometric supervision. While reducing the need for voxel-wise annotations, these approaches still rely on open-vocabulary segmentation models to generate semantic pseudo-labels, making them struggle with generalizing to out-of-distribution scenarios. Besides, they suffer from the computational overhead of dense voxel-wise modeling and volume rendering. DistillNeRF~\cite{DistillNeRF} and OccFeat~\cite{OccFeat} enhance 3D representation learning by distilling volumetric radiance fields with VFMs including DINO~\cite{DINO, DINOv2} and CLIP~\cite{CLIP}. GaussianOcc~\cite{GaussianOcc} improves training efficiency by replacing volume rendering with Gaussian splatting. POP3D~\cite{POP3D} and LangOcc~\cite{LangOcc} align spatial representations with linguistic feature spaces for open-vocabulary predictions.

Our proposed GaussTR diverges from prior works in two key aspects: (1) it adopts sparse, feed-forward 3D Gaussian splatting as an alternative to dense voxel-wise modeling for generalizable representation learning and computational efficiency, and (2) it enables knowledge alignment with foundation models for open-vocabulary 3D semantic occupancy prediction, mitigating the need for explicit labels.

\begin{figure*}
    \centering
    \includegraphics[width=\textwidth]{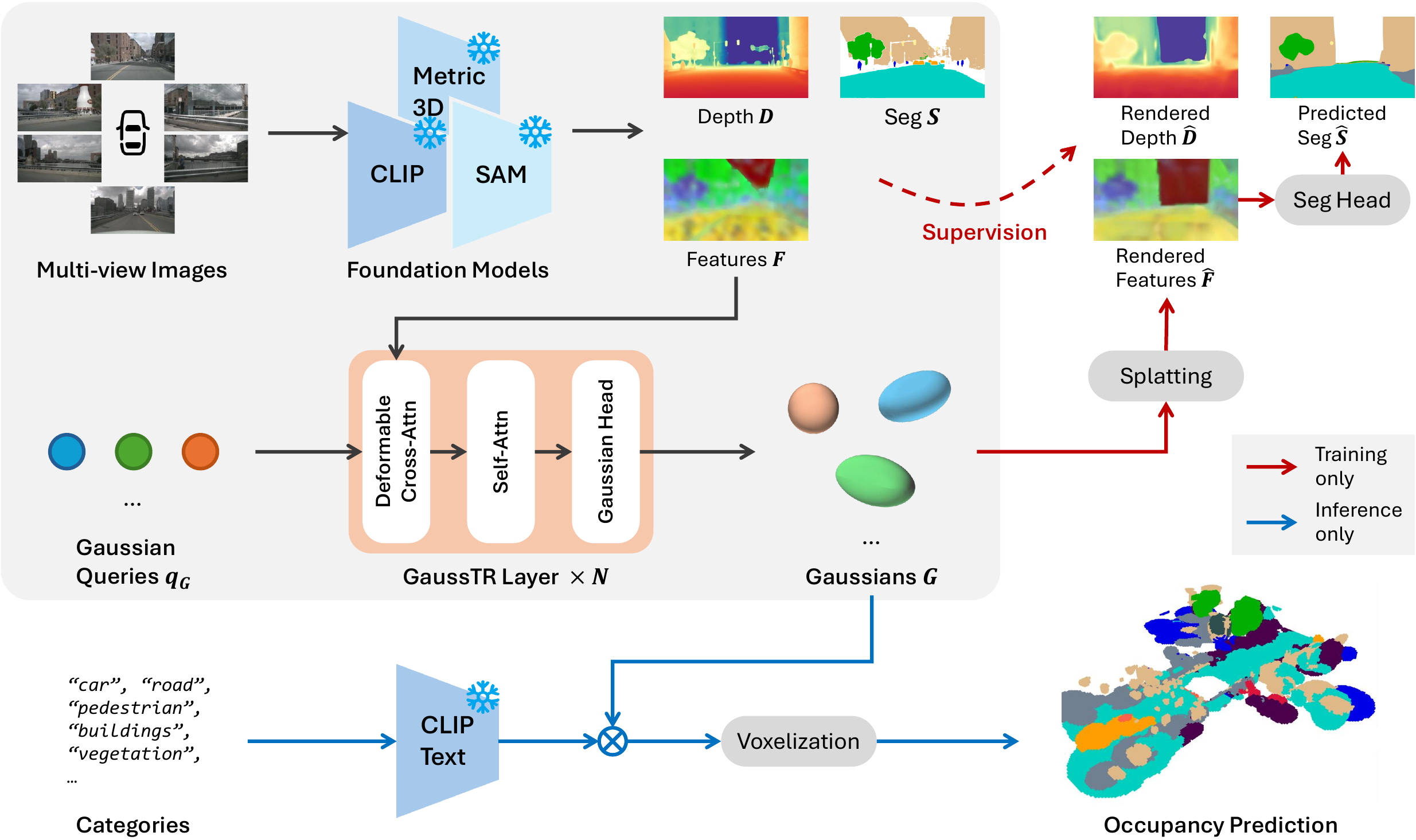}
    \caption{\textbf{Architectural overview of the GaussTR framework.} The GaussTR framework initiates with extracting multi-view features with pre-trained foundation models. A series of Transformer layers then predict sparse sets of Gaussian queries to represent the 3D scene. During the training phase, predicted Gaussians are rendered via differentiable splatting into source 2D views, enforcing alignment with 2D depth and features from foundation models. At inference, Gaussian features are converted into semantic logits by measuring similarity with text-embedded category vectors, followed by voxelization to produce volumetric predictions.}
    \label{fig:method_arch}
\end{figure*}

\subsection{3D Gaussian Splatting}

3D Gaussian Splatting (GS)~\cite{3DGS} has recently emerged as a promising technique for 3D reconstruction, where learnable Gaussians serve as scene representations and improve training and rendering efficiency over voxel-based representations, such as Neural Radiance Fields (NeRF)~\cite{NeRF}. Specifically, 3D GS dynamically adjusts Gaussian properties, including density and covariance, through iterative optimization based on backward gradients. Recent developments in 3D GS have focused on enhancing rendering quality~\cite{MipSplatting, SuGaR}, temporal modeling for dynamic scenes~\cite{4DGaussians}, scene generation~\cite{GaussianDreamer, GaussianDreamerPro}, and feature rendering~\cite{Feature3DGS, FMGS, LangSplat}.

In contrast to conventional 3D GS that optimizes Gaussians independently for each scene, generalizable reconstruction methods~\cite{PixelNeRF} predict Gaussian parameters conditioned on image inputs in a feed-forward manner, enabling the learning of structural priors across multiple scenes. PixelSplat~\cite{PixelSplat} pioneered generalizable 3D GS by sampling Gaussians from predicted probability distributions. Later studies~\cite{GPSGaussian, SplatterImage, Flash3D, MVSplat} typically utilize pre-trained depth estimation networks and predict Gaussian properties in a pixel-wise manner. GeoLRM~\cite{GeoLRM} introduces volumetric occupancy grids for scene modeling and generates Gaussians from them.

Our proposed GaussTR shares conceptual similarities with generalizable 3D GS approaches but differs in its focus on reconstructing high-level 3D scene occupancy rather than RGB images. By leveraging Gaussian splatting, GaussTR enforces feature alignment with foundation models for scalable spatial understanding.

\section{Methodologies}

This section presents a comprehensive elaboration of the GaussTR framework. It commences with outlining the model architecture for feed-forward prediction of Gaussian representations in \cref{sec:gausstr_model}. We then delve into the details of the self-supervised training paradigm in \cref{sec:gausstr_training} and the zero-shot open-vocabulary inference process in \cref{sec:gausstr_inference}. An overview of the GaussTR framework is depicted in \cref{fig:method_arch}.

\subsection{Feedforward Gaussian Splatting}
\label{sec:gausstr_model}

GaussTR first extracts multi-view feature maps $F$ and depth maps $D$ from input images using pre-trained VFMs such as CLIP~\cite{CLIP} and Metric3D V2~\cite{Metric3D, Metric3Dv2}.
We integrate MaskCLIP with FeatUp to enhance the granularity of CLIP features, or Talk2DINO~\cite{Talk2DINO} to extend linguistic alignment for DINO V2~\cite{DINOv2}.
The core of GaussTR revolves around learnable Gaussian queries $q_G \in \mathbb{R}^{N \times C}$, paired with respective pixel positions $\mu_{2D} \in \mathbb{R}^{N \times 2}$ at initialization. Here, $N$ denotes the number of Gaussian queries and $C$ represents the embedding dimension.

Each Gaussian $G$ is parameterized by a set of properties including a 3D center point (mean) $\mu_{3D} \in \mathbb{R}^3$, a 3D covariance matrix $\Sigma$ decomposable into scaling factors $S \in \mathbb{R}^3$ and a rotation quaternion $R \in \mathbb{R}^4$, a density term $\alpha \in [0, 1]$, and a feature vector $f_G \in \mathbb{R}^C$ replacing the spherical harmonics (SH) in conventional GS. Mathematically, this parameterization is expressed as:
\begin{equation}
    G = \{ \mu_{3D}, S, R, \alpha, f_G \}
\end{equation}

The 3D positions $\mu_{3D}$ are initialized via camera-to-world transformation based on depth prediction $D$, camera intrinsics $K$ and extrinsics $E$:
\begin{align}
    \mu_{3D} & = \mathcal{T}_{c2w}(\mu_{2D}, d_G, K, E) \\
             & = E^{-1} K^{-1} (d_G \cdot \mu_{2D})
\end{align}
where $d_G$ is the z-depth coordinate sampled from $D$ at Gaussians' 2D positions $\mu_{2D}$. The initial scale $S_0$ is initialized proportional to $d_G$ to comply with perspective principles, and the initial rotation $R_0$ is initialized orthogonal to camera views derived from $E$.

Through subsequent Transformer decoder layers, GaussTR aggregates multi-scale 2D features from $F$ with deformable attention~\cite{DeformableDETR}, conditioned on projected 2D positions $\mu_{2D}$:
\begin{equation}
    q_G = \text{DeformAttn}(q_G, \mu_{2D}, F)
\end{equation}

Self-attention~\cite{Transformer} across Gaussian queries is then employed for sparse 3D modeling and capturing contextual semantics across the scene, where 3D positional encodings $\text{PE}$ of Gaussian means $\mu_{3D}$ are incorporated into the queries and keys within the attention computation.
\begin{equation}
    q_G = \text{Attn}(q_G + \text{PE}(\mu_{3D}), q_G + \text{PE}(\mu_{3D}), q_G)
\end{equation}

Finally, Gaussian properties are predicted from the Gaussian queries $q_G$ using MLPs within a dedicated Gaussian head:
\begin{equation}
    \{ \Delta \mu_{3D}, \Delta R, \Delta S, \alpha, f_G \} = \text{MLP}(q_G)
\end{equation}
Sigmoid activations and linear transformations are applied to map parameters to appropriate ranges. The Gaussian parameters $\mu_{3D}$, $R$ and $S$ are iteratively refined after each Transformer layer based on predicted deltas $\Delta \mu_{3D}$, $\Delta R$, and $\Delta S$, respectively.

\definecolor{barrier}{RGB}{112, 128, 144}
\definecolor{bicycle}{RGB}{220, 20, 60}
\definecolor{bus}{RGB}{255, 127, 80}
\definecolor{car}{RGB}{255, 158, 0}
\definecolor{construct}{RGB}{233, 150, 70}
\definecolor{motor}{RGB}{255, 61, 99}
\definecolor{pedestrian}{RGB}{0, 0, 230}
\definecolor{traffic}{RGB}{47, 79, 79}
\definecolor{trailer}{RGB}{255, 140, 0}
\definecolor{truck}{RGB}{255, 98, 70}
\definecolor{driveable}{RGB}{0, 207, 191}
\definecolor{other}{RGB}{175, 0, 75}
\definecolor{sidewalk}{RGB}{75, 0, 75}
\definecolor{terrain}{RGB}{112, 180, 60}
\definecolor{manmade}{RGB}{222, 184, 135}
\definecolor{vegetation}{RGB}{0, 175, 0}
\definecolor{others}{RGB}{0, 0, 0}

\begin{table*}[t]
    \centering
    \newcommand{\clsname}[2]{
        \multicolumn{1}{c}{
            \rotatebox{90}{
                \hspace{-6pt}
                \textcolor{#2}{$\blacksquare$} #1
            }}}
    \newcommand{\empa}[1]{\textbf{\underline{#1}}}
    \newcommand{\empb}[1]{\textbf{#1}}
    \setlength{\tabcolsep}{2.07pt}
    \renewcommand\arraystretch{1.05}

    \scalebox{0.96}{
        \begin{tabular}{ l | r>{\columncolor{gray!20}}r | r r r r r r r r r r r r r r r r }
            \toprule
            \textbf{Method}
            & \multicolumn{1}{c}{\textbf{IoU}}
            & \textbf{mIoU}
            & \clsname{barrier}{barrier}
            & \clsname{bicycle}{bicycle}
            & \clsname{bus}{bus}
            & \clsname{car}{car}
            & \clsname{cons. veh.}{construct}
            & \clsname{motorcycle}{motor}
            & \clsname{pedestrian}{pedestrian}
            & \clsname{traffic cone}{traffic}
            & \clsname{trailer}{trailer}
            & \clsname{truck}{truck}
            & \clsname{drive. surf.}{driveable}
            & \clsname{sidewalk}{sidewalk}
            & \clsname{terrain}{terrain}
            & \clsname{manmade}{manmade}
            & \clsname{vegetation}{vegetation} \\
            \midrule
            SelfOcc~\cite{SelfOcc}                 & \empb{45.01}          & 9.30         & 0.15        & 0.66        & 5.46         & 12.54        & 0.00        & 0.80         & 2.10        & 0.00        & 0.00        & 8.25         & \empa{55.49} & \empa{26.30} & \empa{26.54} & 14.22        & 5.60         \\
            OccNeRF~\cite{OccNeRF}                 & 22.81                 & 9.53         & 0.83        & 0.82        & 5.13         & 12.49        & 3.50        & 0.23         & 3.10        & 1.84        & 0.52        & 3.90         & \empb{52.62} & \empb{20.81} & \empb{24.75} & 18.45        & 13.19        \\
            DistillNeRF~\cite{DistillNeRF}         & 29.11                 & 8.93         & 1.35        & 2.08        & 10.21        & 10.09        & 2.56        & 1.98         & 5.54        & \empb{4.62} & \empb{1.43} & 7.90         & 43.02        & 16.86        & 15.02        & 14.06        & \empb{15.06} \\
            GaussianOcc~\cite{GaussianOcc}         & \multicolumn{1}{c}{-} & 9.94         & 1.79        & \empb{5.82} & \empb{14.58} & 13.55        & 1.30        & 2.82         & \empa{7.95} & \empa{9.76} & 0.56        & 9.61         & 44.59        & 20.10        & 17.58        & 8.61         & 10.29        \\
            \midrule
            GaussTR \footnotesize{(FeatUp)}        & \empa{45.19}          & \empb{11.70} & \empb{2.09} & 5.22        & 14.07        & \empb{20.43} & \empb{5.70} & \empb{7.08}  & 5.12        & 3.93        & 0.92        & \empb{13.36} & 39.44        & 15.68        & 22.89        & \empb{21.17} & \empa{21.87} \\
            GaussTR \footnotesize{(Talk2DINO)}     & 44.54                 & \empa{12.27} & \empa{6.50} & \empa{8.54} & \empa{21.77} & \empa{24.27} & \empa{6.26} & \empa{15.48} & \empb{7.94} & 1.86        & \empa{6.10} & \empa{17.16} & 36.98        & 17.21        & 7.16         & \empa{21.18} & 9.99         \\
            \bottomrule
        \end{tabular}
    }

    \caption{
        \textbf{Quantitative performance of self-supervised 3D occupancy methods on the Occ3D-nuScenes~\cite{Occ3D} dataset.}
        For brevity, the IoU scores for the ``others'' and ``other flat'' classes are excluded due to universally zero values across all methods. ``Cons veh." is abbreviated for construction vehicle and ``drive. surf." is for drivable surface. Results are highlighted with the \empa{bold \& underlined} style for the best performance in each column and \empb{bold} for the second-best performance.
    }
    \label{table:occ3d_nus}
\end{table*}

\subsection{VFM-Aligned Self-Supervised Learning}
\label{sec:gausstr_training}

GaussTR bridges 2D visual priors and 3D spatial understanding through Gaussian splatting, thus facilitating self-supervised learning of generalizable 3D representations. The predicted Gaussian representations are rendered to source views and aligned with features and depth maps extracted from foundation models. The Gaussian distribution is described as:
\begin{equation}
    G(x) = e^{-\frac{1}{2}(x)^\mathsf{T} \Sigma^{-1}(x)}
\end{equation}
Here, the covariance matrix $\Sigma$ encodes the spatial extent and orientation of each Gaussian, which consists of scaling factors $S$ and rotation quaternions $R$:
\begin{equation}
    \Sigma = R S S^\mathsf{T} R^\mathsf{T}
\end{equation}

To optimize feature splatting efficiency, Principal Component Analysis (PCA)~\cite{PCA} is applied to reduce the dimensionality of Gaussian features $f_G$. Specifically, we decompose the VFM features $F$ to obtain the principal component $V_k \in \mathbb{R}^{C_R \times C}$, where $C_R$ represents the reduced feature dimensionality. $F$ and $f_G$ are then projected onto the principal components $V_k$.
\begin{align}
    V_k  & = \text{PCA}(F)      \\
    F'   & = F V_k^\mathsf{T}   \\
    f'_G & = f_G V_k^\mathsf{T}
\end{align}

The rendered feature $\hat{F}$ and depth $\hat{D}$ for each pixel are computed by alpha-blending over all Gaussians, replacing the conventional color terms:
\begin{equation}
    \hat{F} = \sum_{i=1}^{N} f'_i \alpha_i \prod_{j=1}^{i-1} (1 - \alpha_j)
\end{equation}

Through Gaussian splatting, GaussTR establishes the consistency of 3D Gaussian representations with 2D VFMs by rendering supervision. The rendered features are supervised by a cosine similarity loss:
\begin{equation}
    \mathcal{L}_{feat} = 1 - \text{cos}(F', \hat{F})
\end{equation}

Depth supervision combines Scale-Invariant Logarithmic (SILog) loss~\cite{SILog} and L1 loss:
\begin{equation}
    \mathcal{L}_{depth} = \mathcal{L}_{SILog}(D, \hat{D}) + \beta \mathcal{L}_{L1}(D, \hat{D})
\end{equation}
\begin{equation}
    \mathcal{L}_{SILog}(D, \hat{D}) = \frac{1}{T} \sum_{i} \delta_i^2 - \frac{1}{T^2} (\sum_i \delta_i)^2
\end{equation}
\begin{equation}
    \delta = \text{log}(D_i) - \text{log}(\hat{D}_i)
\end{equation}
with \(\beta = 0.2\) weighting the terms.

Additionally, segmentation supervision $S$ by Grounded SAM 2~\cite{GroundedSAM} is optionally adopted to refine the semantic boundaries. An auxiliary segmentation head composed of MLPs predicts segmentation maps upon the rendered features $\hat{S} = \text{MLP}(\hat{F})$, which are supervised with a cross-entropy loss:
\begin{equation}
    \mathcal{L}_{seg} = \mathcal{L}_{CE}(S, \hat{S})
\end{equation}

The overall loss is formulated as $\mathcal{L} = \mathcal{L}_{feat} + \mathcal{L}_{depth} + \mathcal{L}_{seg}$, with $\mathcal{L}_{seg}$ optionally applied based on specific training configurations.

\subsection{Open-Vocabulary Occupancy Prediction}
\label{sec:gausstr_inference}

After acquiring Gaussian representations aligned with foundation models, GaussTR enables zero-shot open-vocabulary occupancy prediction leveraging their inherent vision-language consistency. During the inference phase, text prototype embeddings are generated via corresponding text encoder for specified semantic categories (\eg, “car,” “vegetation”), represented as $f_T \in \mathbb{R}^{N_C \times C}$, where $N_C$ denotes the number of categories. Semantic logits for each Gaussian are computed by measuring the similarity between Gaussian features and text features, as given by:
\begin{equation}
    l_G = \text{Softmax}(f_G \cdot f_T^\mathsf{T})
\end{equation}

The resultant logits are then voxelized to produce volumetric occupancy predictions, enabling flexible 3D semantic comprehension across arbitrary categories.  
Notably, when auxiliary segmentation supervision is employed, the categories for mask generation are decoupled from inference categories, while the segmentation head is deactivated during inference, retaining GaussTR's capability for zero-shot, open-vocabulary predictions.
\section{Experiments}

\subsection{Dataset and Metric}

Experiments were conducted on the Occ3D-nuScenes~\cite{Occ3D, nuScenes} dataset, which encompasses 1000 driving scenes with approximately 40,000 frames. The input consists of multi-view images captured from 6 cameras, while the target 3D scene spans a volume of $80 \ \mathrm{m} \times 80 \ \mathrm{m} \times 6.4 \ \mathrm{m}$ with a voxel resolution of $0.4 \ \mathrm{m}$, annotated across 18 semantic classes. Intersection-over-Union (IoU) and mean-IoU (mIoU) metrics are reported for evaluation in line with standard practices.
IoU assesses the binary classification of empty versus occupied voxels, reflecting the accuracy of the geometric reconstruction, whereas the mIoU metric, computed as the average IoU across all classes, provides a comprehensive measurement of semantic understanding and serves as the primary indicator.

\subsection{Implementation Details}

GaussTR was trained for 20 epochs, taking approximately 12 hours on 8 NVIDIA A800 GPUs. The training employs a learning rate of $2 \times 10^{-4}$ and a batch size of $8$. Input image resolution is set at $504 \times 896$. We utilize the ViT-Base~\cite{ViT} model of FeatUp~\cite{FeatUp} and Talk2DINO~\cite{Talk2DINO} as feature backbone and alignment supervision. Metric3D V2~\cite{Metric3Dv2} and Grounded SAM 2~\cite{GroundedSAM} are employed for depth and segmentation supervision, respectively. GaussTR incorporates 300 Gaussian queries per view, amounting to a total of 1800 Gaussians to represent an entire scene. The architecture comprises 3 GaussTR layers with an embedding dimension of $256$.

\subsection{Main Results}

The results of self-supervised occupancy prediction are detailed in \cref{table:occ3d_nus} and \cref{fig:performance_comp}. GaussTR achieves state-of-the-art zero-shot performance of 12.27 mIoU, outperforming previous methods by 2.33 mIoU while reducing training time by 40\%. GaussTR demonstrates performance superiority across diverse foundation models including FeatUp~\cite{FeatUp} and Talk2DINO~\cite{Talk2DINO}. These findings validate the scalability and generalization of sparse Gaussian-based 3D modeling and foundation model alignment for self-supervised spatial understanding.

Specifically, GaussTR particularly excels in object-centric categories, \eg, cars, trucks and manmade structures (\ie, buildings). However, it struggles with small objects such as traffic cones, as well as flat surfaces like roads and sidewalks. This discrepancy arises because while Gaussian-based 3D representations well fit object-centric modeling, they encounter challenges in capturing less prominent objects with sparse representations, and flat surfaces are prone to occlusion during the splatting-based alignment process.

\subsection{Ablation Studies}

\paragraph{Ablation on Geometric Alignment.} To evaluate the impact of geometric alignment, we isolate the contributions of Metric3D V2~\cite{Metric3Dv2} and FeatUp~\cite{FeatUp}, as shown in \cref{table:ablat_geom}. It is revealed that training with feature alignment alone fails to converge, probably attributed to the optimization dilemma in jointly optimizing spatial positions and features for Gaussian splatting, as also observed in PixelSplat~\cite{PixelSplat}.
In contrast, Metric3D alone yields a solid geometric performance of 43.87 IoU, while the integration of feature alignment can further elevate the IoU score, suggesting their synergistic effect.

\begin{table}[ht]
    \centering
    \begin{tabular}{ c c | r r }
        \toprule
        Metric3D   & FeatUp     & \multicolumn{1}{c}{IoU}            & \multicolumn{1}{c}{mIoU} \\
        \midrule
                   & \checkmark & \multicolumn{2}{c}{not convergent}                            \\
        \checkmark &            & 43.87                              & \multicolumn{1}{c}{-}    \\
        \checkmark & \checkmark & \textbf{45.19}                     & \textbf{11.70}           \\
        \bottomrule
    \end{tabular}

    \caption{\textbf{Ablation on geometric alignment.}}
    \label{table:ablat_geom}
\end{table}


\paragraph{Ablation on Feature Alignment.} 
We analyze the generalizability of GaussTR by comparing foundation models for feature alignment: \textbf{MaskCLIP~\cite{MaskCLIP}} adapts the original CLIP to reproduce language-aligned feature maps; \textbf{FeatUp~\cite{FeatUp}} is employed upon MaskCLIP to compensate for spatial granularity through feature upsampling; \textbf{Talk2DINO~\cite{Talk2DINO}} aligns DINO V2~\cite{DINOv2} visual features with text embeddings for vision-language reasoning.

As presented in \cref{table:ablat_feat}, the introduction of FeatUp with MaskCLIP boosts the performance by 0.93 mIoU, attributed to its enhancement of spatial granularity for CLIP features. Talk2DINO improves mIoU by 0.57, showcasing the superiority of DINO visual representations. Interestingly, while auxiliary segmentation supervision benefits FeatUp alignment by 0.94 mIoU by refining semantic boundaries, it significantly degrades the performance when applied to Talk2DINO, which we hypothesize reflects the optimization conflict when representations are already robust.

\begin{table}[ht]
    \centering
    \setlength{\tabcolsep}{2.5pt}
    \begin{tabular}{ c c c c | r r }
        \toprule
        MaskCLIP   & FeatUp     & Talk2DINO  & Aux. Seg.  & \multicolumn{1}{c}{IoU} & \multicolumn{1}{c}{mIoU} \\
        \midrule
        \checkmark &            &            &            & 42.45                   & 9.83                     \\
        \checkmark & \checkmark &            &            & 44.43                   & 10.76                    \\
        \checkmark & \checkmark &            & \checkmark & \textbf{45.19}          & \textbf{11.70}           \\
                   &            & \checkmark & \checkmark & 44.89                   & 11.34                    \\
                   &            & \checkmark &            & \textbf{44.54}          & \textbf{12.27}           \\
        \bottomrule
    \end{tabular}

    \caption{\textbf{Ablation on feature alignment.}}
    \label{table:ablat_feat}
\end{table}



\begin{figure*}
    \centering
    \includegraphics[width=\textwidth]{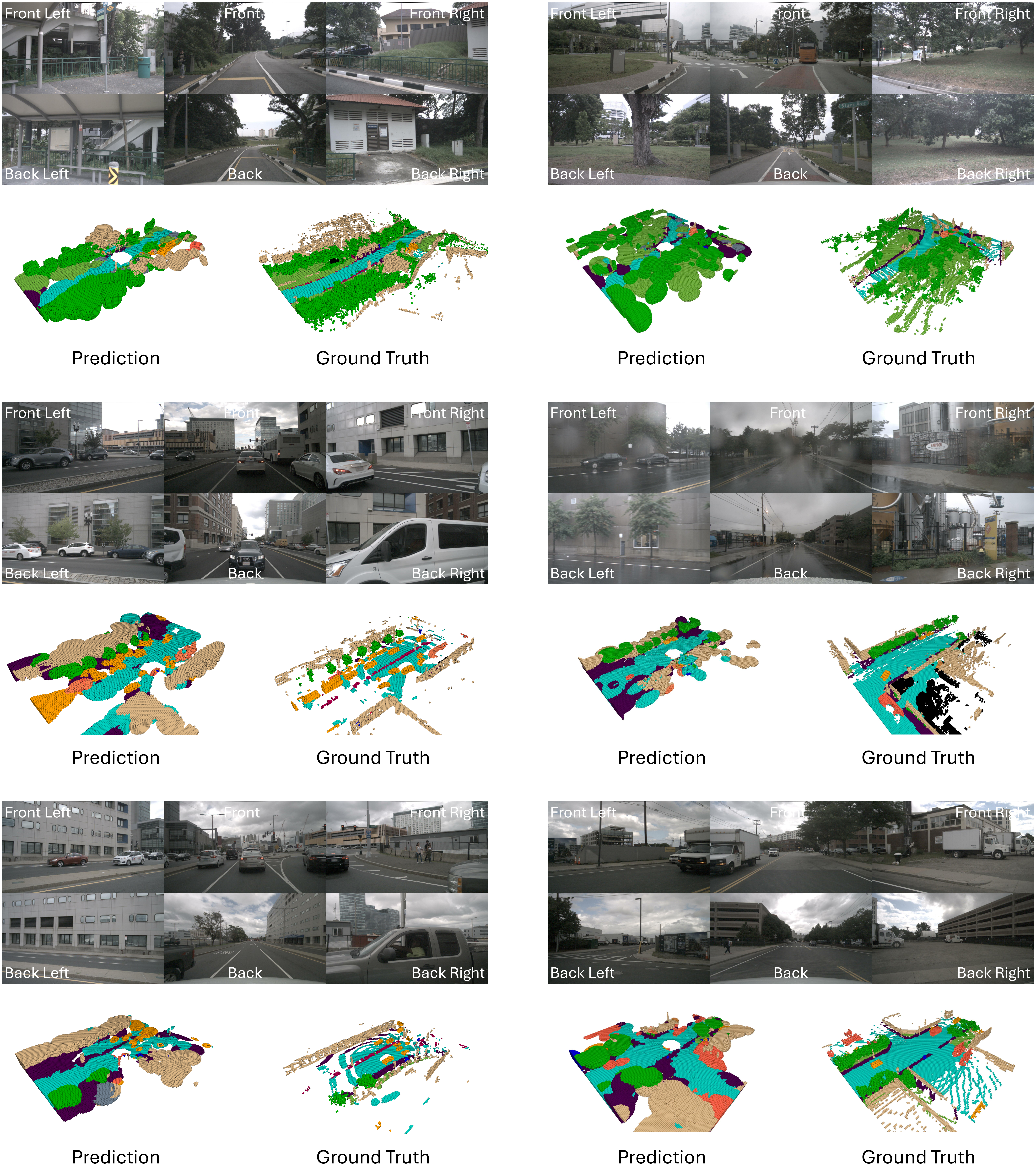}
    \caption{\textbf{Qualitative visualizations of GaussTR on Occ3D-nuScenes~\cite{Occ3D}.} GaussTR consistently produces both a coherent global scene structures and fine-grained local details, offering a comprehensive understanding of the environment. Notably, it excels at modeling object-centric categories, such as cars and buildings.}
    \label{fig:vis1}
\end{figure*}

\begin{figure*}
    \centering
    \includegraphics[width=\textwidth]{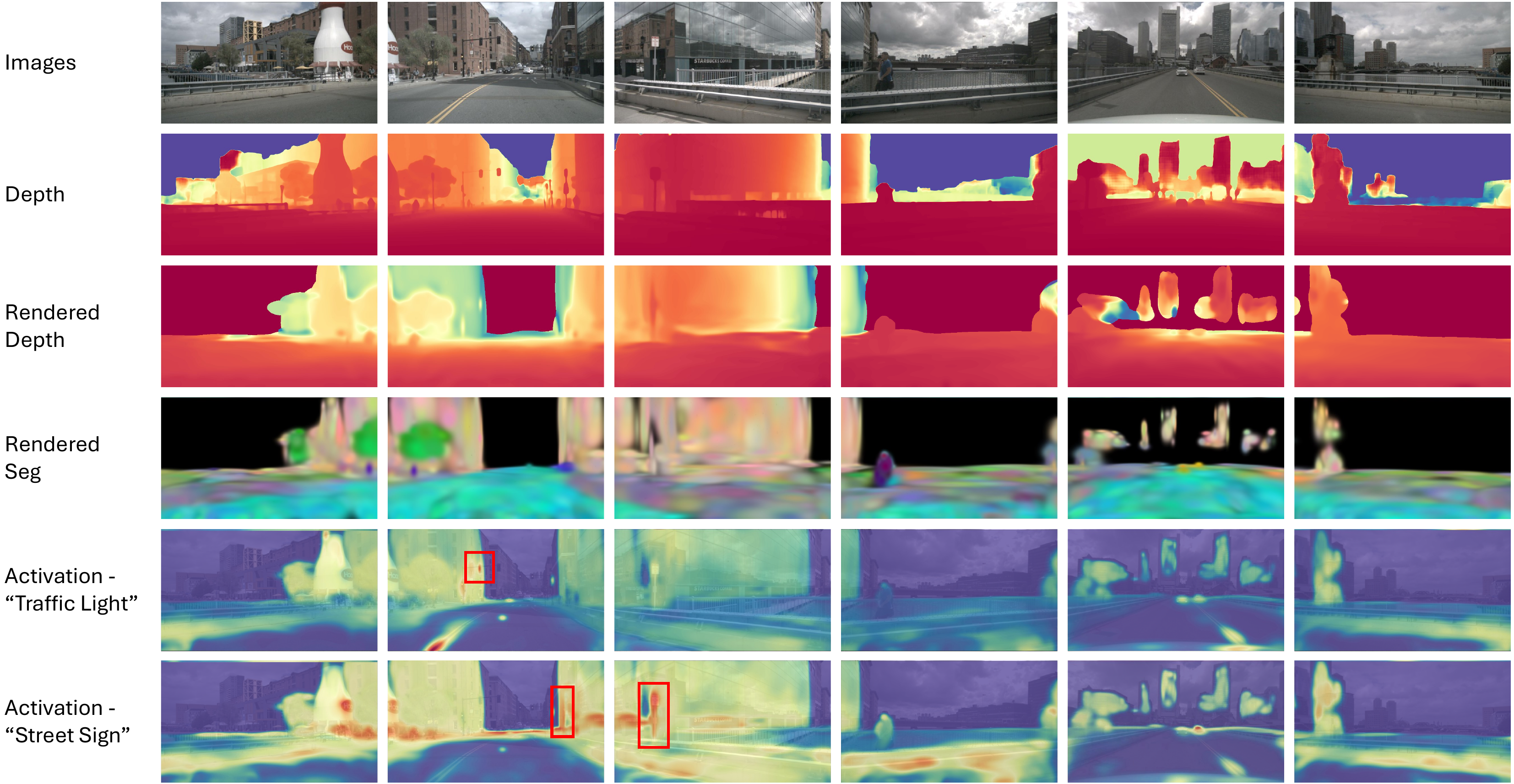}
    \caption{\textbf{Visualizations of rendered views.} The figure illustrates the rendered depth and segmentation maps of Gaussian predictions derived from camera views. Moreover, activation maps for novel categories are visualized, highlighted in red boxes.}
    \label{fig:vis2}
\end{figure*}

\paragraph{Ablation on Number of Gaussian Queries.} \cref{table:ablat_num_queries} analyzes performance across varying numbers of Gaussian queries, foundation models, and auxiliary segmentation. Results indicate that increasing queries generally enhances performance, while the higher-performing model with higher performance can leverage more Gaussian representations to achieve optimal results.
For FeatUp, performance almost plateaus at 200 in the absence of auxiliary segmentation but continues improving until 300 Gaussians when segmentation supervision is applied. In contrast, Talk2DINO sustains performance gains even at 400 queries, suggesting its capacity for richer representation spaces.

Notably, FeatUp’s performance degrades after 300 queries, likely attributed to attention dilution in 3D modeling, where excessive Gaussian queries overwhelm the model's capacity to focus on critical spatial interactions among them. This observation also highlights a potential direction for future research, which is to optimize the sequential modeling of 3D scenes to maintain performance scalability.

\begin{table}[ht]
    \centering
    \setlength{\tabcolsep}{4.5pt}
    \begin{tabular}{ c | c c | c c | c c }
        \toprule
        \multirow{3}{*}{\#Queries}
        & \multicolumn{2}{c}{\textbf{FeatUp}}
        & \multicolumn{2}{|c}{\textbf{FeatUp}}
        & \multicolumn{2}{|c}{\textbf{Talk2DINO}} \\
        & \multicolumn{2}{c}{\footnotesize{w/o aux. seg.}}
        & \multicolumn{2}{|c}{\footnotesize{w/ aux. seg.}}
        & \multicolumn{2}{|c}{\footnotesize{w/o aux. seg.}} \\
        & \multicolumn{1}{c}{IoU}  & \multicolumn{1}{c}{mIoU}
        & \multicolumn{1}{|c}{IoU} & \multicolumn{1}{c}{mIoU}
        & \multicolumn{1}{|c}{IoU} & \multicolumn{1}{c}{mIoU} \\
        \midrule
        100       & 42.06          & 10.02          & 41.84          & 10.53          & 41.83          & 10.81          \\
        200       & 44.25          & 10.73          & 44.32          & 11.46          & 43.92          & 11.68          \\
        300       & \textbf{44.73} & \textbf{10.76} & \textbf{45.19} & \textbf{11.70} & 44.54          & 12.27          \\
        400       & 43.91          & 10.54          & 44.65          & 11.52          & \textbf{45.66} & \textbf{12.45} \\
        \bottomrule
    \end{tabular}

    \caption{\textbf{Ablation on number of queries.}}
    \label{table:ablat_num_queries}
\end{table}


\subsection{Visualizations}

\cref{fig:vis1} illustrates the qualitative visualizations of GaussTR on Occ3D-nuScenes. GaussTR conducts robust scene understanding by jointly capturing global geometric coherence and fine-grained object details. These results showcase the advantages of GaussTR in delineating spatial relationships across global scenes while providing sparse, object-centric representations. It also sheds light on the distribution of predicted Gaussians, offering insight into the internal mechanisms of the sparse Gaussian-based 3D modeling.
GaussTR demonstrates remarkable precision in recognizing and modeling object categories such as cars, pedestrians, and buildings, while revealing limitations in reconstructing flat surfaces, such as roads, due to occlusion-induced ambiguities during splatting. These observations align with quantitative results in \cref{table:occ3d_nus}. The superior performance of GaussTR can be attributed to its alignment with foundation models and the intrinsic sparsity of its representations, which facilitate efficient scene interpretation.

In \cref{fig:vis2}, we further analyze GaussTR’s cross-modal consistency by visualizing rendered 2D depth and segmentation maps of Gaussian predictions. To improve interpretability, we apply color perturbations to the semantic maps to highlight the distribution of individual Gaussians and reveal how they collectively reconstruct the scene layout.
Additionally, GaussTR exhibits impressive generalization capability to novel and scarce categories, such as traffic lights and street signs. Owing to its alignment with visual-language models, GaussTR can seamlessly adapt to these categories, generating prominent activations in corresponding regions and further validating its versatility.

\section{Conclusion}

This paper introduces GaussTR, a Gaussian-based Transformer framework that aligns with foundation models to advance self-supervised 3D spatial understanding. GaussTR represents scenes as sparse Gaussian queries through feed-forward Transformer prediction, further aligned with pre-trained VFMs to learn general 3D representations via differentiable splatting. This foundation model alignment facilitates open-vocabulary semantic occupancy prediction without explicit annotations, alleviating the scalability and generalization limitations of prior methods. Empirical experiments showcase GaussTR's state-of-the-art performance of 12.27 mIoU with improved efficiency. Overall, GaussTR pioneers a novel paradigm that leverages sparse Gaussian representations and foundation model alignment. We envision GaussTR as a promising pathway towards scalable and generalizable 3D spatial intelligence for autonomous driving, robotics, and beyond.

\paragraph{Acknowledgement.}
This work was partially supported by the National Natural Science Foundation of China (NSFC) under Grant No. 62376102.

{
    \small
    \bibliographystyle{ieeenat_fullname}
    \bibliography{main}

\begin{thebibliography}{63}
\providecommand{\natexlab}[1]{#1}
\providecommand{\url}[1]{\texttt{#1}}
\expandafter\ifx\csname urlstyle\endcsname\relax
  \providecommand{\doi}[1]{doi: #1}\else
  \providecommand{\doi}{doi: \begingroup \urlstyle{rm}\Url}\fi

\bibitem[Abdi and Williams(2010)]{PCA}
Herv{\'e} Abdi and Lynne~J Williams.
\newblock Principal component analysis.
\newblock \emph{Wiley interdisciplinary reviews: computational statistics}, 2\penalty0 (4):\penalty0 433--459, 2010.

\bibitem[Barsellotti et~al.(2024)Barsellotti, Bianchi, Messina, Carrara, Cornia, Baraldi, Falchi, and Cucchiara]{Talk2DINO}
Luca Barsellotti, Lorenzo Bianchi, Nicola Messina, Fabio Carrara, Marcella Cornia, Lorenzo Baraldi, Fabrizio Falchi, and Rita Cucchiara.
\newblock Talking to dino: Bridging self-supervised vision backbones with language for open-vocabulary segmentation.
\newblock \emph{arXiv preprint arXiv:2411.19331}, 2024.

\bibitem[Boeder et~al.(2024)Boeder, Gigengack, and Risse]{LangOcc}
Simon Boeder, Fabian Gigengack, and Benjamin Risse.
\newblock Langocc: Self-supervised open vocabulary occupancy estimation via volume rendering.
\newblock \emph{arXiv preprint arXiv:2407.17310}, 2024.

\bibitem[Caesar et~al.(2020)Caesar, Bankiti, Lang, Vora, Liong, Xu, Krishnan, Pan, Baldan, and Beijbom]{nuScenes}
Holger Caesar, Varun Bankiti, Alex~H Lang, Sourabh Vora, Venice~Erin Liong, Qiang Xu, Anush Krishnan, Yu Pan, Giancarlo Baldan, and Oscar Beijbom.
\newblock nuscenes: A multimodal dataset for autonomous driving.
\newblock In \emph{CVPR}, pages 11621--11631, 2020.

\bibitem[Cao and De~Charette(2022)]{MonoScene}
Anh-Quan Cao and Raoul De~Charette.
\newblock Monoscene: Monocular 3d semantic scene completion.
\newblock In \emph{CVPR}, pages 3991--4001, 2022.

\bibitem[Cao and de~Charette(2023)]{SceneRF}
Anh-Quan Cao and Raoul de Charette.
\newblock Scenerf: Self-supervised monocular 3d scene reconstruction with radiance fields.
\newblock In \emph{ICCV}, pages 9387--9398, 2023.

\bibitem[Caron et~al.(2021)Caron, Touvron, Misra, J{\'e}gou, Mairal, Bojanowski, and Joulin]{DINO}
Mathilde Caron, Hugo Touvron, Ishan Misra, Herv{\'e} J{\'e}gou, Julien Mairal, Piotr Bojanowski, and Armand Joulin.
\newblock Emerging properties in self-supervised vision transformers.
\newblock In \emph{ICCV}, pages 9650--9660, 2021.

\bibitem[Charatan et~al.(2024)Charatan, Li, Tagliasacchi, and Sitzmann]{PixelSplat}
David Charatan, Sizhe~Lester Li, Andrea Tagliasacchi, and Vincent Sitzmann.
\newblock pixelsplat: 3d gaussian splats from image pairs for scalable generalizable 3d reconstruction.
\newblock In \emph{CVPR}, pages 19457--19467, 2024.

\bibitem[Chen et~al.(2025)Chen, Xu, Zheng, Zhuang, Pollefeys, Geiger, Cham, and Cai]{MVSplat}
Yuedong Chen, Haofei Xu, Chuanxia Zheng, Bohan Zhuang, Marc Pollefeys, Andreas Geiger, Tat-Jen Cham, and Jianfei Cai.
\newblock Mvsplat: Efficient 3d gaussian splatting from sparse multi-view images.
\newblock In \emph{ECCV}, pages 370--386, 2025.

\bibitem[Cheng et~al.(2024)Cheng, Jiang, Chen, Liao, Zhang, Liu, and Wang]{BVT}
Tianheng Cheng, Haoyi Jiang, Shaoyu Chen, Bencheng Liao, Qian Zhang, Wenyu Liu, and Xinggang Wang.
\newblock Learning accurate monocular 3d voxel representation via bilateral voxel transformer.
\newblock \emph{Image Vis. Comput.}, 150:\penalty0 105237, 2024.

\bibitem[Dong et~al.(2023)Dong, Bao, Zheng, Zhang, Chen, Yang, Zeng, Zhang, Yuan, Chen, et~al.]{MaskCLIP}
Xiaoyi Dong, Jianmin Bao, Yinglin Zheng, Ting Zhang, Dongdong Chen, Hao Yang, Ming Zeng, Weiming Zhang, Lu Yuan, Dong Chen, et~al.
\newblock Maskclip: Masked self-distillation advances contrastive language-image pretraining.
\newblock In \emph{CVPR}, pages 10995--11005, 2023.

\bibitem[Dosovitskiy et~al.(2021)Dosovitskiy, Beyer, Kolesnikov, Weissenborn, Zhai, Unterthiner, Dehghani, Minderer, Heigold, Gelly, Uszkoreit, and Houlsby]{ViT}
Alexey Dosovitskiy, Lucas Beyer, Alexander Kolesnikov, Dirk Weissenborn, Xiaohua Zhai, Thomas Unterthiner, Mostafa Dehghani, Matthias Minderer, Georg Heigold, Sylvain Gelly, Jakob Uszkoreit, and Neil Houlsby.
\newblock An image is worth 16x16 words: Transformers for image recognition at scale.
\newblock In \emph{ICLR}, 2021.

\bibitem[Eigen et~al.(2014)Eigen, Puhrsch, and Fergus]{SILog}
David Eigen, Christian Puhrsch, and Rob Fergus.
\newblock Depth map prediction from a single image using a multi-scale deep network.
\newblock \emph{NeurIPS}, 27, 2014.

\bibitem[Fang et~al.(2023)Fang, Wang, Xie, Sun, Wu, Wang, Huang, Wang, and Cao]{EVA}
Yuxin Fang, Wen Wang, Binhui Xie, Quan Sun, Ledell Wu, Xinggang Wang, Tiejun Huang, Xinlong Wang, and Yue Cao.
\newblock Eva: Exploring the limits of masked visual representation learning at scale.
\newblock In \emph{CVPR}, pages 19358--19369, 2023.

\bibitem[Fang et~al.(2024)Fang, Sun, Wang, Huang, Wang, and Cao]{EVA02}
Yuxin Fang, Quan Sun, Xinggang Wang, Tiejun Huang, Xinlong Wang, and Yue Cao.
\newblock Eva-02: A visual representation for neon genesis.
\newblock \emph{Image and Vis. Comput.}, 149:\penalty0 105171, 2024.

\bibitem[Fu et~al.(2024)Fu, Hamilton, Brandt, Feldman, Zhang, and Freeman]{FeatUp}
Stephanie Fu, Mark Hamilton, Laura Brandt, Axel Feldman, Zhoutong Zhang, and William~T Freeman.
\newblock Featup: A model-agnostic framework for features at any resolution.
\newblock \emph{arXiv preprint arXiv:2403.10516}, 2024.

\bibitem[Gan et~al.(2024)Gan, Liu, Xu, Mo, and Yokoya]{GaussianOcc}
Wanshui Gan, Fang Liu, Hongbin Xu, Ningkai Mo, and Naoto Yokoya.
\newblock Gaussianocc: Fully self-supervised and efficient 3d occupancy estimation with gaussian splatting.
\newblock \emph{arXiv preprint arXiv:2408.11447}, 2024.

\bibitem[Godard et~al.(2019)Godard, Mac~Aodha, Firman, and Brostow]{MonoDepth2}
Cl{\'e}ment Godard, Oisin Mac~Aodha, Michael Firman, and Gabriel~J Brostow.
\newblock Digging into self-supervised monocular depth estimation.
\newblock In \emph{ICCV}, pages 3828--3838, 2019.

\bibitem[Gu{\'e}don and Lepetit(2024)]{SuGaR}
Antoine Gu{\'e}don and Vincent Lepetit.
\newblock Sugar: Surface-aligned gaussian splatting for efficient 3d mesh reconstruction and high-quality mesh rendering.
\newblock In \emph{CVPR}, pages 5354--5363, 2024.

\bibitem[Hayler et~al.(2024)Hayler, Wimbauer, Muhle, Rupprecht, and Cremers]{S4C}
Adrian Hayler, Felix Wimbauer, Dominik Muhle, Christian Rupprecht, and Daniel Cremers.
\newblock S4c: Self-supervised semantic scene completion with neural fields.
\newblock In \emph{3DV}, pages 409--420, 2024.

\bibitem[Hu et~al.(2024)Hu, Yin, Zhang, Cai, Long, Chen, Wang, Yu, Shen, and Shen]{Metric3Dv2}
Mu Hu, Wei Yin, Chi Zhang, Zhipeng Cai, Xiaoxiao Long, Hao Chen, Kaixuan Wang, Gang Yu, Chunhua Shen, and Shaojie Shen.
\newblock Metric3d v2: A versatile monocular geometric foundation model for zero-shot metric depth and surface normal estimation.
\newblock \emph{arXiv preprint arXiv:2404.15506}, 2024.

\bibitem[Huang et~al.(2021)Huang, Huang, Zhu, Ye, and Du]{BEVDet}
Junjie Huang, Guan Huang, Zheng Zhu, Yun Ye, and Dalong Du.
\newblock Bevdet: High-performance multi-camera 3d object detection in bird-eye-view.
\newblock \emph{arXiv preprint arXiv:2112.11790}, 2021.

\bibitem[Huang et~al.(2023)Huang, Zheng, Zhang, Zhou, and Lu]{TPVFormer}
Yuanhui Huang, Wenzhao Zheng, Yunpeng Zhang, Jie Zhou, and Jiwen Lu.
\newblock Tri-perspective view for vision-based 3d semantic occupancy prediction.
\newblock In \emph{CVPR}, pages 9223--9232, 2023.

\bibitem[Huang et~al.(2024{\natexlab{a}})Huang, Zheng, Zhang, Zhou, and Lu]{SelfOcc}
Yuanhui Huang, Wenzhao Zheng, Borui Zhang, Jie Zhou, and Jiwen Lu.
\newblock Selfocc: Self-supervised vision-based 3d occupancy prediction.
\newblock In \emph{CVPR}, pages 19946--19956, 2024{\natexlab{a}}.

\bibitem[Huang et~al.(2024{\natexlab{b}})Huang, Zheng, Zhang, Zhou, and Lu]{GaussianFormer}
Yuanhui Huang, Wenzhao Zheng, Yunpeng Zhang, Jie Zhou, and Jiwen Lu.
\newblock Gaussianformer: Scene as gaussians for vision-based 3d semantic occupancy prediction.
\newblock In \emph{ECCV}, 2024{\natexlab{b}}.

\bibitem[Jiang et~al.(2024)Jiang, Cheng, Gao, Zhang, Lin, Liu, and Wang]{Symphonies}
Haoyi Jiang, Tianheng Cheng, Naiyu Gao, Haoyang Zhang, Tianwei Lin, Wenyu Liu, and Xinggang Wang.
\newblock Symphonize 3d semantic scene completion with contextual instance queries.
\newblock In \emph{CVPR}, pages 20258--20267, 2024.

\bibitem[Kerbl et~al.(2023)Kerbl, Kopanas, Leimk{\"u}hler, and Drettakis]{3DGS}
Bernhard Kerbl, Georgios Kopanas, Thomas Leimk{\"u}hler, and George Drettakis.
\newblock 3d gaussian splatting for real-time radiance field rendering.
\newblock \emph{ACM TOG}, 42\penalty0 (4):\penalty0 139--1, 2023.

\bibitem[Kirillov et~al.(2023)Kirillov, Mintun, Ravi, Mao, Rolland, Gustafson, Xiao, Whitehead, Berg, Lo, et~al.]{SAM}
Alexander Kirillov, Eric Mintun, Nikhila Ravi, Hanzi Mao, Chloe Rolland, Laura Gustafson, Tete Xiao, Spencer Whitehead, Alexander~C Berg, Wan-Yen Lo, et~al.
\newblock Segment anything.
\newblock In \emph{ICCV}, pages 4015--4026, 2023.

\bibitem[Li et~al.(2023)Li, Yu, Choy, Xiao, Alvarez, Fidler, Feng, and Anandkumar]{VoxFormer}
Yiming Li, Zhiding Yu, Christopher Choy, Chaowei Xiao, Jose~M Alvarez, Sanja Fidler, Chen Feng, and Anima Anandkumar.
\newblock Voxformer: Sparse voxel transformer for camera-based 3d semantic scene completion.
\newblock In \emph{CVPR}, pages 9087--9098, 2023.

\bibitem[Li et~al.(2022)Li, Wang, Li, Xie, Sima, Lu, Qiao, and Dai]{BEVFormer}
Zhiqi Li, Wenhai Wang, Hongyang Li, Enze Xie, Chonghao Sima, Tong Lu, Yu Qiao, and Jifeng Dai.
\newblock Bevformer: Learning bird’s-eye-view representation from multi-camera images via spatiotemporal transformers.
\newblock In \emph{ECCV}, pages 1--18, 2022.

\bibitem[Liu et~al.(2024)Liu, Wang, Chen, Yang, Zeng, Chen, and Wang]{SparseOcc}
Haisong Liu, Haiguang Wang, Yang Chen, Zetong Yang, Jia Zeng, Li Chen, and Limin Wang.
\newblock Fully sparse 3d panoptic occupancy prediction.
\newblock In \emph{ECCV}, 2024.

\bibitem[Ma et~al.(2024{\natexlab{a}})Ma, Chen, Huang, Xu, Luo, Xu, Gu, Ai, and Wang]{Cam4DOcc}
Junyi Ma, Xieyuanli Chen, Jiawei Huang, Jingyi Xu, Zhen Luo, Jintao Xu, Weihao Gu, Rui Ai, and Hesheng Wang.
\newblock Cam4docc: Benchmark for camera-only 4d occupancy forecasting in autonomous driving applications.
\newblock In \emph{CVPR}, pages 21486--21495, 2024{\natexlab{a}}.

\bibitem[Ma et~al.(2024{\natexlab{b}})Ma, Tan, Qu, Ma, Zhang, and Xie]{COTR}
Qihang Ma, Xin Tan, Yanyun Qu, Lizhuang Ma, Zhizhong Zhang, and Yuan Xie.
\newblock Cotr: Compact occupancy transformer for vision-based 3d occupancy prediction.
\newblock In \emph{CVPR}, pages 19936--19945, 2024{\natexlab{b}}.

\bibitem[Mildenhall et~al.(2021)Mildenhall, Srinivasan, Tancik, Barron, Ramamoorthi, and Ng]{NeRF}
Ben Mildenhall, Pratul~P Srinivasan, Matthew Tancik, Jonathan~T Barron, Ravi Ramamoorthi, and Ren Ng.
\newblock Nerf: Representing scenes as neural radiance fields for view synthesis.
\newblock \emph{Communications of the ACM}, 65\penalty0 (1):\penalty0 99--106, 2021.

\bibitem[Oquab et~al.(2023)Oquab, Darcet, Moutakanni, Vo, Szafraniec, Khalidov, Fernandez, Haziza, Massa, El-Nouby, et~al.]{DINOv2}
Maxime Oquab, Timoth{\'e}e Darcet, Th{\'e}o Moutakanni, Huy Vo, Marc Szafraniec, Vasil Khalidov, Pierre Fernandez, Daniel Haziza, Francisco Massa, Alaaeldin El-Nouby, et~al.
\newblock Dinov2: Learning robust visual features without supervision.
\newblock \emph{arXiv preprint arXiv:2304.07193}, 2023.

\bibitem[Pan et~al.(2024)Pan, Liu, Zhang, Huang, Li, Xie, Wang, Liu, and Zhang]{RenderOcc}
Mingjie Pan, Jiaming Liu, Renrui Zhang, Peixiang Huang, Xiaoqi Li, Hongwei Xie, Bing Wang, Li Liu, and Shanghang Zhang.
\newblock Renderocc: Vision-centric 3d occupancy prediction with 2d rendering supervision.
\newblock In \emph{ICRA}, pages 12404--12411, 2024.

\bibitem[Qin et~al.(2024)Qin, Li, Zhou, Wang, and Pfister]{LangSplat}
Minghan Qin, Wanhua Li, Jiawei Zhou, Haoqian Wang, and Hanspeter Pfister.
\newblock Langsplat: 3d language gaussian splatting.
\newblock In \emph{CVPR}, pages 20051--20060, 2024.

\bibitem[Radford et~al.(2021)Radford, Kim, Hallacy, Ramesh, Goh, Agarwal, Sastry, Askell, Mishkin, Clark, et~al.]{CLIP}
Alec Radford, Jong~Wook Kim, Chris Hallacy, Aditya Ramesh, Gabriel Goh, Sandhini Agarwal, Girish Sastry, Amanda Askell, Pamela Mishkin, Jack Clark, et~al.
\newblock Learning transferable visual models from natural language supervision.
\newblock In \emph{ICML}, pages 8748--8763, 2021.

\bibitem[Ren et~al.(2024)Ren, Liu, Zeng, Lin, Li, Cao, Chen, Huang, Chen, Yan, et~al.]{GroundedSAM}
Tianhe Ren, Shilong Liu, Ailing Zeng, Jing Lin, Kunchang Li, He Cao, Jiayu Chen, Xinyu Huang, Yukang Chen, Feng Yan, et~al.
\newblock Grounded sam: Assembling open-world models for diverse visual tasks.
\newblock \emph{arXiv preprint arXiv:2401.14159}, 2024.

\bibitem[Shi et~al.(2024)Shi, Cheng, Zhang, Liu, and Wang]{OSP}
Yiang Shi, Tianheng Cheng, Qian Zhang, Wenyu Liu, and Xinggang Wang.
\newblock Occupancy as set of points.
\newblock In \emph{ECCV}, 2024.

\bibitem[Sirko-Galouchenko et~al.(2024)Sirko-Galouchenko, Boulch, Gidaris, Bursuc, Vobecky, P{\'e}rez, and Marlet]{OccFeat}
Sophia Sirko-Galouchenko, Alexandre Boulch, Spyros Gidaris, Andrei Bursuc, Antonin Vobecky, Patrick P{\'e}rez, and Renaud Marlet.
\newblock Occfeat: Self-supervised occupancy feature prediction for pretraining bev segmentation networks.
\newblock In \emph{CVPR}, pages 4493--4503, 2024.

\bibitem[Song et~al.(2017)Song, Yu, Zeng, Chang, Savva, and Funkhouser]{SSCNet}
Shuran Song, Fisher Yu, Andy Zeng, Angel~X Chang, Manolis Savva, and Thomas Funkhouser.
\newblock Semantic scene completion from a single depth image.
\newblock In \emph{CVPR}, pages 1746--1754, 2017.

\bibitem[Szymanowicz et~al.(2024{\natexlab{a}})Szymanowicz, Insafutdinov, Zheng, Campbell, Henriques, Rupprecht, and Vedaldi]{Flash3D}
Stanislaw Szymanowicz, Eldar Insafutdinov, Chuanxia Zheng, Dylan Campbell, Jo{\~a}o~F Henriques, Christian Rupprecht, and Andrea Vedaldi.
\newblock Flash3d: Feed-forward generalisable 3d scene reconstruction from a single image.
\newblock \emph{arXiv preprint arXiv:2406.04343}, 2024{\natexlab{a}}.

\bibitem[Szymanowicz et~al.(2024{\natexlab{b}})Szymanowicz, Rupprecht, and Vedaldi]{SplatterImage}
Stanislaw Szymanowicz, Chrisitian Rupprecht, and Andrea Vedaldi.
\newblock Splatter image: Ultra-fast single-view 3d reconstruction.
\newblock In \emph{CVPR}, pages 10208--10217, 2024{\natexlab{b}}.

\bibitem[Tian et~al.(2024)Tian, Jiang, Yun, Mao, Yang, Wang, Wang, and Zhao]{Occ3D}
Xiaoyu Tian, Tao Jiang, Longfei Yun, Yucheng Mao, Huitong Yang, Yue Wang, Yilun Wang, and Hang Zhao.
\newblock Occ3d: A large-scale 3d occupancy prediction benchmark for autonomous driving.
\newblock \emph{NeurIPS}, 36, 2024.

\bibitem[Vaswani et~al.(2017)Vaswani, Shazeer, Parmar, Uszkoreit, Jones, Gomez, Kaiser, and Polosukhin]{Transformer}
Ashish Vaswani, Noam Shazeer, Niki Parmar, Jakob Uszkoreit, Llion Jones, Aidan~N. Gomez, Lukasz Kaiser, and Illia Polosukhin.
\newblock Attention is all you need.
\newblock \emph{NeurIPS}, pages 5998--6008, 2017.

\bibitem[Vobecky et~al.(2024)Vobecky, Sim{\'e}oni, Hurych, Gidaris, Bursuc, P{\'e}rez, and Sivic]{POP3D}
Antonin Vobecky, Oriane Sim{\'e}oni, David Hurych, Spyridon Gidaris, Andrei Bursuc, Patrick P{\'e}rez, and Josef Sivic.
\newblock Pop-3d: Open-vocabulary 3d occupancy prediction from images.
\newblock \emph{NeurIPS}, 36, 2024.

\bibitem[Wang et~al.(2024)Wang, Kim, Yang, Yu, Ivanovic, Waslander, Wang, Fidler, Pavone, and Karkus]{DistillNeRF}
Letian Wang, Seung~Wook Kim, Jiawei Yang, Cunjun Yu, Boris Ivanovic, Steven~L Waslander, Yue Wang, Sanja Fidler, Marco Pavone, and Peter Karkus.
\newblock Distillnerf: Perceiving 3d scenes from single-glance images by distilling neural fields and foundation model features.
\newblock \emph{NeurIPS}, 2024.

\bibitem[Wimbauer et~al.(2023)Wimbauer, Yang, Rupprecht, and Cremers]{BTS}
Felix Wimbauer, Nan Yang, Christian Rupprecht, and Daniel Cremers.
\newblock Behind the scenes: Density fields for single view reconstruction.
\newblock In \emph{CVPR}, pages 9076--9086, 2023.

\bibitem[Wu et~al.(2024)Wu, Yi, Fang, Xie, Zhang, Wei, Liu, Tian, and Wang]{4DGaussians}
Guanjun Wu, Taoran Yi, Jiemin Fang, Lingxi Xie, Xiaopeng Zhang, Wei Wei, Wenyu Liu, Qi Tian, and Xinggang Wang.
\newblock 4d gaussian splatting for real-time dynamic scene rendering.
\newblock In \emph{CVPR}, pages 20310--20320, 2024.

\bibitem[Yi et~al.(2024{\natexlab{a}})Yi, Fang, Wang, Wu, Xie, Zhang, Liu, Tian, and Wang]{GaussianDreamer}
Taoran Yi, Jiemin Fang, Junjie Wang, Guanjun Wu, Lingxi Xie, Xiaopeng Zhang, Wenyu Liu, Qi Tian, and Xinggang Wang.
\newblock Gaussiandreamer: Fast generation from text to 3d gaussians by bridging 2d and 3d diffusion models.
\newblock In \emph{CVPR}, pages 6796--6807, 2024{\natexlab{a}}.

\bibitem[Yi et~al.(2024{\natexlab{b}})Yi, Fang, Zhou, Wang, Wu, Xie, Zhang, Liu, Wang, and Tian]{GaussianDreamerPro}
Taoran Yi, Jiemin Fang, Zanwei Zhou, Junjie Wang, Guanjun Wu, Lingxi Xie, Xiaopeng Zhang, Wenyu Liu, Xinggang Wang, and Qi Tian.
\newblock Gaussiandreamerpro: Text to manipulable 3d gaussians with highly enhanced quality.
\newblock \emph{arXiv preprint arXiv:2406.18462}, 2024{\natexlab{b}}.

\bibitem[Yin et~al.(2023)Yin, Zhang, Chen, Cai, Yu, Wang, Chen, and Shen]{Metric3D}
Wei Yin, Chi Zhang, Hao Chen, Zhipeng Cai, Gang Yu, Kaixuan Wang, Xiaozhi Chen, and Chunhua Shen.
\newblock Metric3d: Towards zero-shot metric 3d prediction from a single image.
\newblock In \emph{ICCV}, pages 9043--9053, 2023.

\bibitem[Yu et~al.(2021)Yu, Ye, Tancik, and Kanazawa]{PixelNeRF}
Alex Yu, Vickie Ye, Matthew Tancik, and Angjoo Kanazawa.
\newblock pixelnerf: Neural radiance fields from one or few images.
\newblock In \emph{CVPR}, pages 4578--4587, 2021.

\bibitem[Yu et~al.(2023)Yu, Shu, Deng, Lu, Liu, Yu, Yang, Li, and Chen]{FlashOcc}
Zichen Yu, Changyong Shu, Jiajun Deng, Kangjie Lu, Zongdai Liu, Jiangyong Yu, Dawei Yang, Hui Li, and Yan Chen.
\newblock Flashocc: Fast and memory-efficient occupancy prediction via channel-to-height plugin.
\newblock \emph{arXiv preprint arXiv:2311.12058}, 2023.

\bibitem[Yu et~al.(2024)Yu, Chen, Huang, Sattler, and Geiger]{MipSplatting}
Zehao Yu, Anpei Chen, Binbin Huang, Torsten Sattler, and Andreas Geiger.
\newblock Mip-splatting: Alias-free 3d gaussian splatting.
\newblock In \emph{CVPR}, pages 19447--19456, 2024.

\bibitem[Zhang et~al.(2023{\natexlab{a}})Zhang, Yan, Wei, Li, Liu, Tang, Duan, and Lu]{OccNeRF}
Chubin Zhang, Juncheng Yan, Yi Wei, Jiaxin Li, Li Liu, Yansong Tang, Yueqi Duan, and Jiwen Lu.
\newblock Occnerf: Self-supervised multi-camera occupancy prediction with neural radiance fields.
\newblock \emph{arXiv preprint arXiv:2312.09243}, 2023{\natexlab{a}}.

\bibitem[Zhang et~al.(2024)Zhang, Song, Wei, Chen, Lu, and Tang]{GeoLRM}
Chubin Zhang, Hongliang Song, Yi Wei, Yu Chen, Jiwen Lu, and Yansong Tang.
\newblock Geolrm: Geometry-aware large reconstruction model for high-quality 3d gaussian generation.
\newblock \emph{arXiv preprint arXiv:2406.15333}, 2024.

\bibitem[Zhang et~al.(2023{\natexlab{b}})Zhang, Zhu, and Du]{OccFormer}
Yunpeng Zhang, Zheng Zhu, and Dalong Du.
\newblock Occformer: Dual-path transformer for vision-based 3d semantic occupancy prediction.
\newblock In \emph{ICCV}, pages 9433--9443, 2023{\natexlab{b}}.

\bibitem[Zheng et~al.(2024)Zheng, Zhou, Shao, Liu, Zhang, Nie, and Liu]{GPSGaussian}
Shunyuan Zheng, Boyao Zhou, Ruizhi Shao, Boning Liu, Shengping Zhang, Liqiang Nie, and Yebin Liu.
\newblock Gps-gaussian: Generalizable pixel-wise 3d gaussian splatting for real-time human novel view synthesis.
\newblock In \emph{CVPR}, pages 19680--19690, 2024.

\bibitem[Zhou et~al.(2024)Zhou, Chang, Jiang, Fan, Zhu, Xu, Chari, You, Wang, and Kadambi]{Feature3DGS}
Shijie Zhou, Haoran Chang, Sicheng Jiang, Zhiwen Fan, Zehao Zhu, Dejia Xu, Pradyumna Chari, Suya You, Zhangyang Wang, and Achuta Kadambi.
\newblock Feature 3dgs: Supercharging 3d gaussian splatting to enable distilled feature fields.
\newblock In \emph{CVPR}, pages 21676--21685, 2024.

\bibitem[Zhu et~al.(2020)Zhu, Su, Lu, Li, Wang, and Dai]{DeformableDETR}
Xizhou Zhu, Weijie Su, Lewei Lu, Bin Li, Xiaogang Wang, and Jifeng Dai.
\newblock Deformable detr: Deformable transformers for end-to-end object detection.
\newblock In \emph{ICLR}, 2020.

\bibitem[Zuo et~al.(2024)Zuo, Samangouei, Zhou, Di, and Li]{FMGS}
Xingxing Zuo, Pouya Samangouei, Yunwen Zhou, Yan Di, and Mingyang Li.
\newblock Fmgs: Foundation model embedded 3d gaussian splatting for holistic 3d scene understanding.
\newblock \emph{IJCV}, pages 1--17, 2024.

\end{thebibliography}
}


\end{document}